\documentclass{article}
\pdfoutput=1

\usepackage{arxiv}

\usepackage[utf8]{inputenc} % allow utf-8 input
\usepackage[T1]{fontenc}    % use 8-bit T1 fonts
\usepackage{hyperref}       % hyperlinks
\usepackage{url}            % simple URL typesetting
\usepackage{booktabs}       % professional-quality tables
\usepackage{amsfonts}       % blackboard math symbols
\usepackage{nicefrac}       % compact symbols for 1/2, etc.
\usepackage{microtype}      % microtypography
\usepackage{lipsum}		% Can be removed after putting your text content
\usepackage{graphicx}
\usepackage{natbib}
\usepackage{doi}

\title{On the Impact of Temporal Representations on Metaphor Detection}

\author{ {\href{https://orcid.org/my-orcid?orcid=0000-0002-0822-0345}{\includegraphics[scale=0.06]{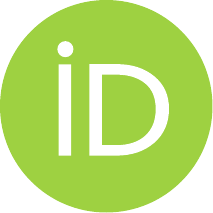}\hspace{1mm}Giorgio Ottolina}} \\%\thanks{Use footnote for providing further
	Systems and Communication\\
	University of Milano-Bicocca\\
	Milan, Italy \\
	\texttt{g.ottolina1@campus.unimib.it} \\
	\And
	\href{https://orcid.org/0000-0002-1801-5118}{\includegraphics[scale=0.06]{orcid.pdf}\hspace{1mm}Matteo Palmonari} \\
	Systems and Communication\\
	University of Milano-Bicocca\\
	Milan, Italy \\
	\texttt{matteo.palmonari@unimib.it} \\
	\And
	\href{https://orcid.org/0000-0002-7867-6612}{\includegraphics[scale=0.06]{orcid.pdf}\hspace{1mm}Mehwish Alam} \\
	FIZ Karlsruhe - Leibniz Institute for Information Infrastructure\\
	Karlsruhe Institute of Technology\\
	Karlsruhe, Germany \\
	\texttt{mehwish.alam@fiz-karlsruhe.de} \\
	\And
    {\hspace{1mm}Manuel Vimercati} \\
	Systems and Communication\\
	University of Milano-Bicocca\\
	Milan, Italy \\
	\texttt{manuel.vimercati@unimib.it} \\
}

\renewcommand{\shorttitle}{\textit{arXiv} Template}

\hypersetup{
pdftitle={A template for the arxiv style},
pdfsubject={q-bio.NC, q-bio.QM},
pdfauthor={David S.~Hippocampus, Elias D.~Striatum},
pdfkeywords={First keyword, Second keyword, More},
}

\begin{document}
\maketitle

\begin{abstract}
State-of-the-art approaches for metaphor detection compare their literal - or core - meaning and their contextual meaning using metaphor classifiers based on neural networks. However, metaphorical expressions evolve over time due to various reasons, such as cultural and societal impact. Metaphorical expressions are known to co-evolve with language and literal word meanings, and even drive, to some extent, this evolution. This poses the question of whether different, possibly time-specific, representations of literal meanings may impact the metaphor detection task. To the best of our knowledge, this is the first study that examines the metaphor detection task with a detailed exploratory analysis where different temporal and static word embeddings are used to account for different representations of literal meanings. Our experimental analysis is based on three popular benchmarks used for metaphor detection and word embeddings extracted from different corpora and temporally aligned using different state-of-the-art approaches. The results suggest that the usage of different static word embedding methods does impact the metaphor detection task and some temporal word embeddings slightly outperform static methods. However, the results also suggest that temporal word embeddings may provide representations of the core meaning of the metaphor even too close to their contextual meaning, thus confusing the classifier. Overall, the interaction between temporal language evolution and metaphor detection appears tiny in the benchmark datasets used in our experiments. This suggests that future work for the computational analysis of this important linguistic phenomenon should first start by creating a new dataset where this interaction is better represented.
\end{abstract}

% keywords can be removed

\section{Introduction}
\label{introduction}

%%%%%%%%%%%%%%%%%%%
%Mehwish
%%%%%%%%%%%%%%%%%%%

%\begin{itemize}
%    \item What is figurative language? more specifically metaphorical expressions.
%    \item Does it create any challenges for machine understanding? Yes it does.
%    \item What kind of challenges? Here give a reference and possibly an example but optional.
%    \item Discuss changes in literal text over time along with the causes of these changes, give an example and a reference.
%    \item Similar changes for metaphors. Concrete example? 
%    \item to solve this problem we are making an initial set of experiments....
%    \item now is the time to summarize what we did... the points above motivates our work.
%\end{itemize}

Accounting for figurative language is one of the key challenges in Natural Language Processing (NLP)~\cite{RecuperoABGT19,shutova2}. Figurative language often contains metaphorical expressions which map one concept from a source domain to another concept in a target domain. For instance, in the sentence ``\textit{The  wheels of Stalin’s regime were well-oiled and already turning}", a political system (target concept) is viewed in terms of a mechanism (source concept) that can function, break, have wheels, etc. This association allows us to transfer knowledge from the domain of \textit{mechanical engineering} to that of \textit{politics}. Therefore, political systems are thought about in terms of mechanisms, leading to multiple metaphorical expressions. The phenomenon of source-target domain mapping was first introduced by George Lakoff known as Conceptual Metaphor Theory~\cite{lakoff}. Due to previously defined characteristics, the presence of metaphorical expression in text causes misinterpretation in the algorithms such as machine translation or sentiment analysis \cite{Mohammad}.   

Recent studies addressing the metaphor detection problem are based on machine learning and exploit word embeddings~\cite{shutova2,leong-etal-2020-report}, often relying on pre-trained models as linguistic resources. The key intuition is to recognize that words are used in a context that is different from their usual context. 
In the previous example, the term ``\textit{wheels}" is collocated, in the sentence, close to ``\textit{Stalin}" and ``\textit{regime}", thus defining a context different from the contexts in which it usually appears, i.e., in the domain of mechanical engineering. Most of the recent approaches have therefore combined non-contextual and contextual word embeddings to provide signals for this comparison~\cite{mao-etal-2019-end,swarnkar-singh-2018-di}. For example, \cite{mao-etal-2019-end,inproceedings,Word2vec} combine non-contextual GloVe embeddings \cite{glove} with contextual ELMo embeddings \cite{contextual2} within a BiLSTM neural network for sequence labeling. GloVe embeddings account for literal word meanings, while ELMo embeddings account for contextual word meanings.

An important linguistic phenomenon that is not considered in state-of-the-art methods for metaphor detection is language evolution. The trait of the evolution of meaning over time is also shared by metaphorical expressions, which can be due to various reasons such as cultural and societal impact. Metaphorical expressions are known to co-evolve with language and literal word meanings drive this evolution to some extent \cite{inbook,hickey_2003}. This leads to the question of whether different, possibly time-specific, representations of literal meanings impact the task of metaphor detection. In conclusion, if metaphor detection approaches tend to compare a sentence-specific and a literal meaning, we must be aware that literal meaning as accounted for in static word embeddings 1) depends on the corpus (the reference linguistic resource) and method used to train the embeddings, and 2) evolves over time.  

To this end, this empirical study focuses on analyzing the impact of different word embeddings accounting for literal word meaning on the task of metaphor detection. Special attention should be paid to possible interactions between metaphor detection and time-specific (non–contextual) word representations used to account for literal meanings at different times. The empirical study discussed in this paper aims to make a first step towards addressing the co-evolution of metaphors and language evolution which is known to be an important factor in language evolution itself~\cite{inbook,hickey_2003}. 

The methodology adopted in this study consists of the following protocol. First, a state-of-the-art Recurrent Neural Networks (RNN)-based model~\cite{gao1}, which uses static word embeddings to account for literal word meaning, is selected for metaphor detection. This model performs metaphor detection as a sequence classification task where each word occurrence is labeled as either a metaphor usage or a literal usage. Second, three widely used benchmark datasets are selected to evaluate the performance of the models. Third, the RNN-based model is fed with literal meaning vectors obtained from different (non-contextual) word embeddings including temporal word embeddings computed for different decades and aligned with state-of-the-art alignment methods, such as \textit{Procrustes} \cite{Procrustes} and \textit{Compass}~\cite{CADE} (first referred to as Temporal Word Embeddings with a Compass~\cite{TWEC} - TWEC).

The experimental results indicate that different word embeddings impact the metaphor detection task and some temporal word embeddings slightly outperform classic methods on some performance measures. These quantitative results are then explained with the help of a qualitative analysis of the predictions made by the models. 
An example that illustrates our findings is given in the following figurative sentence coming from a state-of-the-art dataset (see Section~\ref{datasets}), which has been mistakenly classified as literal by a model exploiting an atemporal embedding, and correctly detected as metaphorical by the same model exploiting a temporal word embedding: \textit{``The \underline{virus attacked} Argonne National Laboratory outside Chicago starting at 11.54 pm EST Wednesday and throughout the night}". If we investigate the ten nearest neighbors of ``\textit{\underline{virus}}", in a temporal embedding we find words such as ``\textit{infection", ``respiratory"} and \textit{``organism}", while in the atemporal one we find, for example, ``\textit{malware"} and \textit{``spyware}", which diverge from the original literal core meaning and are related to a modern connotation of the word. When exploiting the temporal word embedding, the model is able to understand that the ``\textit{\underline{virus}}" in this sentence is a computer virus, and therefore is used metaphorically along with the verb ``\textit{\underline{attacked}}". However, the analysis provided in this paper also suggests that temporal word embeddings may provide representations of words' core meaning too close to their metaphorical meaning, thus confusing the classifier. 

The paper is organized as follows: Section~\ref{sota} discusses the related work about metaphor detection as well as temporal word embeddings. Section~\ref{methodology} discusses the methodology which has been followed, while Section~\ref{experiments} shows the experimental results of the paper. Finally, Section~\ref{discussion} concludes the paper.

\section{State of the Art}
\label{sota}

\label{related_work}

This section discusses the state-of-the-art approaches for metaphor detection and the studies related to temporal language evolution. 

\subsection{Early Approaches for Metaphor Detection}
In~\cite{wilks}, the author proposes an approach for metaphor detection based on preferential semantics, which affirms that metaphors are ``a violation of semantic constraints put by verbs onto their arguments". In~\cite{fass}, the author proposes an approach for processing metonyms as well as metaphors that take into account the distinction between literalness, metonymy, metaphoricity, and anomaly. This work uses hand-coded patterns for testing sentences containing metonymic relations. The drawback of this approach was that the interpretations were always context-dependent. In \cite{peters}, the authors use WordNet hierarchy to group senses and to find hyponymy relations. If two words are not included in the same synset and/or in hierarchically related synsets, then they are most likely part of a metaphorical phrase. CorMet~\cite{CorMet} was the first system to automatically discover source-target domain mappings. A survey on these approaches has been given in~\cite{shutova2}.

\subsection{Neural Network Based Approaches for Metaphor Detection}
Numerous approaches based on BiLSTM take advantage of both contextualized and pre-trained embeddings in the classification layer~\cite{mao-etal-2019-end,swarnkar-singh-2018-di}. In particular, the \textit{Di-LSTM} Contrast system~\cite{swarnkar-singh-2018-di} encodes the left and right side context of a target word through forward and backward LSTMs. The classification is based on a concatenation of the target word representation and its difference with the encoded context. \cite{mao-etal-2019-end} combined GloVe and BiLSTM hidden states for sequence labeling. Some of the most recent systems fine-tune pre-trained contextual language models such as BERT~\cite{transformers2} and RoBERTa~\cite{roberta}. For example, \cite{dankers-etal-2020-neighbourly} fine-tuned a BERT model, fed with a discourse fragment as input. Hierarchical attention computes both token and sentence level attention after the encoded layers, leading to better results.
A more detailed discussion on methods for metaphor detection is given in this dedicated survey~\cite{RaiC20}. Another recent approach~\cite{li-etal-2021-label} uses the hierarchical contextualized representation to extract more information from both sentence-level and discourse-level. For our study we tested the approaches \cite{gao1} for two main reasons: they explicitly model the interaction between literal and contextual meaning (and thus they support the replacement of embeddings accounting for literal word meaning with different corpus and time-specific embeddings) and they achieved state-of-the-art performance on several metaphor detection datasets when we started our study.

\subsection{Temporal Language Evolution}

According to \cite{hickey_2003}, theories often come as a formalization of metaphors, which 
``\textit{can populate history with new objects and kinds, and provide both access to interesting new worlds and great field-internal success}".
Based on the observations that language is always changing 
\cite{five2009language,complexity,book}, linguists have formulated different theories and models searching for rules and regularities in semantic change, such as the ``\textit{Diachronic Prototype Semantics}" \cite{Geeraerts1997-GEEDPS,book1}, the ``\textit{Invited Inference Theory of Semantic Change}" \cite{traugott_dasher_2001}, and ``\textit{semantic change based on metaphor and metonymy}" \cite{article}.

Historically, much of the theoretical work on semantic shifts has been devoted to documenting and categorizing various types of semantic shifts \cite{nla.cat-vn2374050,Stern1975-STEMAC}.
Semantic shifts are separated into two important classes: ``\textit{linguistic drifts}" (slow and steady changes in core meaning of words) and ``\textit{cultural shifts}" (changes in associations of a given word determined by cultural influences). In \cite{inproceedings}, the authors showed that distributional models capture cultural shifts, like the word ``\textit{sleep}" acquiring more negative connotations related to the sleep disorders domain, when comparing its 1960s contexts with its 1990s contexts. Researchers studying semantic change from a computational point of view have empirically shown the existence of this distinction \cite{Hamilton}.

Diachronic corpora provide empirical resources for semantic change analysis.
The availability of large corpora  enabled the development of new methodologies for the study of lexical-semantic shifts within general linguistics \cite{traugott_dasher_2001}. A key assumption is that changes in a word’s collocational patterns reflect changes in word meaning, thus providing a usage-based account of semantic shifts. Semantic changes are often reflected in large corpora that can be sliced into time-specific chunks (e.g., texts coming from a same decade), which account for changes in the contexts of a word that is affected by the shift.
Most recent approaches to studying diachronic semantic change are based on temporal word embeddings. These approaches are based on  (1) slicing a corpus into time-specific slices (e.g., one slice per decade), and (2) generating slice-specific representations by solving the cross-slice alignment problem~\cite{Hamilton}. Other novel approaches include \cite{giulianelli-etal-2020-analysing} those based on contextualized word embeddings and \cite{tsakalidis-liakata-2020-sequential} those based on sequential modeling. The current study uses temporal word embeddings for decade-specific slices generated with two alignment methods: HistWords (SGNS) embeddings aligned with the \textit{Procustes} method ~\cite{Hamilton,Diachronic} and CADE word embeddings aligned with the \textit{Compass} method \cite{CADE,TWEC}. HistWords embeddings are used in this study because they are used in the key studies about semantic change and are available as pre-trained embeddings; CADE embeddings are used because they achieved state-of-the-art performance on different tasks at the time this study was started \cite{TWEC}.

\section{Methodology} \label{methodology}

This paper considers the metaphor detection task in its more general settings: the task consists in \textit{detecting all the occurrences of words used metaphorically in an input sentence, independently from  their POS tags}.
To analyze the effect of different word embeddings (i.e., temporal or static) on the metaphor detection task, the metaphor detection approaches proposed in~\cite{mao-etal-2019-end} were selected. These approaches use both (non-contextual) word embeddings and contextual word embeddings within a neural network with a final classification layer. 

The goal is to train and test different instances of the same architecture with different (non-contextual) word embeddings that account for literal meanings in the metaphor detection algorithms,
to verify whether using word representations derived from different linguistic resources leads to different classification outcomes.

\subsection{Metaphor Detection Approach}
\label{metaphor-detection}
In \cite{mao-etal-2019-end}, two end-to-end metaphor identification models for detecting metaphors are proposed, both performing better
than the previous state-of-the-art baseline \cite{gao1}.
The two proposed state-of-the-art models are: \textit{(i) Recurrent Neural Network Hidden GloVe (RNN HG)}, based on the interaction between literal and contextual word meanings; \textit{(ii) Multi-Head Context Attention (RNN MHCA)}, based on multi-head context attention. It was observed that the RNN HG and RNN MHCA models achieve comparable results on state-of-the-art metaphor detection datasets. After some preliminary experiments, RNN HG was found to be more suitable to our concerns, since the static embedding is explicitly digested by the network and compared with the contextual embedding. 

\begin{figure}[htp]
        \centering
        \includegraphics[scale=0.3]{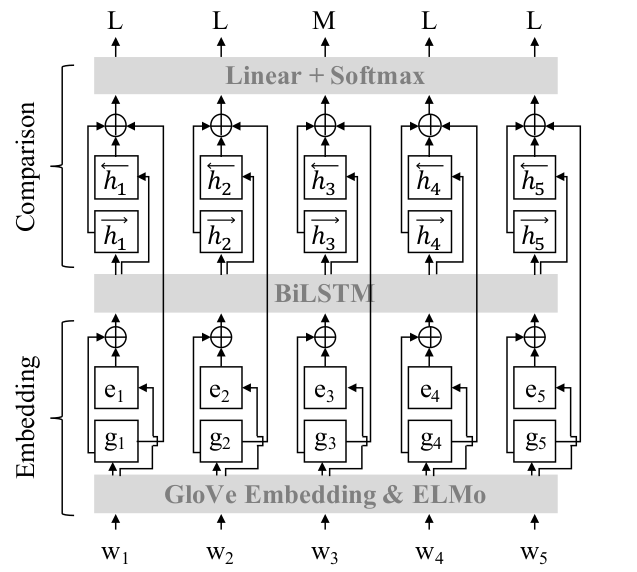}
        \caption{RNN HG model architecture based on MIP procedure. \label{im5}}
    \end{figure}

\paragraph{RNN HG.} 
Figure~\ref{im5} shows the overall architecture of RNN HG as described in the original study.
The RNN HG model can be represented through the following equations:

$$
{t}=f_{B i L S T M}\left(\left[g_{t} ; e_{t}\right], \vec{h}_{t-1}, \stackrel{\leftarrow}{h}_{t+1}\right)$$

$$
p\left(\hat{y}_{t} \mid h_{t}, g_{t}\right)=\sigma\left(w^{\top}\left(h_{t} ; g_{t}\right)+b\right)$$

where: \begin{math} {h_{t}} \end{math} is the hidden state; \begin{math} {g_{t}} \end{math} is the input GloVe \cite{glove} literal representation; \begin{math}{e_{t}} \end{math} is the input ELMo \cite{contextual2} representation; \begin{math} w \end{math} is trained parameters; \begin{math} \sigma \end{math} is the softmax function; \begin{math} \hat{y} \end{math} is the the probability of a label prediction for a target word at position \begin{math} t \end{math}; \begin{math} {t} \end{math} is the target word.

In the original architecture/model GloVe embeddings serve as literal (non-contextual) representations of a word (\begin{math} {g_{t}} \end{math}) and are concatenated with the representation from the hidden layer (\begin{math} {h_{t}} \end{math}) of a BiLSTM. These embeddings, located in two different encoding spaces, are concatenated feeding the BiLSTM network and  fulfilling the MIP~\cite{discourse} requirement.
The literal and contextual representations then get compared in the so-called comparison stage. This last step consists of a softmax function \begin{math} \sigma \end{math}, which computes the probability of a label prediction \begin{math} \hat{y} \end{math} for a target word at position \textit t, conditioned on both its contextual and literal meaning representations. 

In RNN HG, both static and contextual representations are used to account for the differences (or the similarities) between them. Also, from the equation it can be easily seen that the vector of the static representation can be modified without touching the model. The different static representations used in this paper are explained in section \ref{temporal}.

\subsection{Word Embeddings}
\label{temporal}

Word embeddings providing literal word meanings in the RNN HG network used in our experiments are of two main kinds: temporal word embeddings and static word embeddings. Both kinds of embeddings are trained with different approaches and corpora to account for several variables that are expected to have an impact on the final representations (especially: corpus, word embedding algorithm, and alignment method). 

\subsubsection{Temporal Word Embeddings}

\textit{HistWords - SGNS}\footnote{\url{https://stanford.io/3txN0Hd}} provides a set of pre-trained temporal word embeddings generated using the Skip-gram variant of Word2vec \cite{Word2vec} trained with negative sampling (also referred to as SGNS) on different sliced diachronic corpora~\cite{hamilton_diachronic_2016,Diachronic}. Decade-specific embeddings obtained from the same corpus are aligned using the Procrustes method, one of the most used in the literature. It solves the task of aligning two sets of points in high dimensions (which has many applications in NLP), through the joint estimation of an orthogonal matrix and a permutation matrix~\cite{Procrustes}. A stochastic algorithm is proposed to minimize the cost function on large-scale problems. In our study we consider HistWords embeddings obtained from different corpora. \textbf{CoHa Word SGNS} (1900-2010) is a set of eleven decade-specific models covering the time span 1900-2010 (with ``1900" we refer to the ``1900-1910" slice); they are trained on a genre-balanced subset of the Corpus of Historical American English (CoHa)~\cite{DVN/8SRSYK_2015}, the largest structured corpus of historical English, which contains more than 400 million words and text published between 1820 and 2000s. 
\textbf{CoHa Lemma SGNS} (1900-2010) is a set of eleven decade-specific models trained on CoHa after applying lemmatization. \textbf{NGrams English All} and \textbf{NGrams English Fiction} (1900-2000) are two sets of ten decade-specific models each trained on a subset of Google N-Grams that considers, respectively, all genres or fiction only. Observe that we consider a total of 42 models (each one containing decade-specific word embeddings) based on Procrustes alignment. In the experiments, we may filter out some models that do not achieve the best performance for space limitations.

\textit{CADE - Compass Aligned Embeddings} are temporal word embeddings trained with Word2vec \cite{Word2vec2} and aligned with the Compass method~\cite{TWEC,CADE}, which can be summarized as follows. Word2vec is trained over the entire corpus (all the slices). One of the two-weight matrices obtained after this step is used as a \textit{compass} when training Word2vec again on each slice: the compass matrix is frozen, while the other matrix is initialized and trained again over each slice, thus obtaining slice-specific word embeddings (the word embeddings referring to the slice period). We use CADE with the CBOW architecture as in the original paper \cite{TWEC}, thus using the target matrix as a compass and the weights in the context matrix as final embeddings. CADE embeddings used in the study are trained using the code and implementation details available online\footnote{\url{https://github.com/vinid/cade}}. To obtain embeddings as comparable as possible to CoHa Word SGNS, we trained CADE embeddings using the CoHa corpus.\footnote{Unfortunately, the genre-balanced subset of CoHA used in HistWords could not be retrieved to train the embeddings on the very same data. Also, the CBOW architecture has been used because it is reported to generate temporal word embeddings of better quality with the compass \cite{TWEC}} \textbf{CoHa Word CBOW} (1900-2010) is the set of eleven slice-specific models trained with this approach.    

\subsubsection{Static Word Embeddings}
Three static word embeddings obtained from as many corpora are also considered to account for the impact of the corpus and embedding algorithm on metaphor detection.  

\textbf{Common Crawl GloVe} is the model that is based on the embeddings used in the original RNN HG network. These embeddings are trained using the \textbf{Common Crawl}~\footnote{\url{https://commoncrawl.org/}} corpus, which is expected to contain relatively recent content extracted from the web. \textbf{Wikipedia CBOW} consists of the embeddings trained over the English Wikipedia using Word2vec with the CBOW architecture. It accounts for a relatively recent text covering encyclopedic knowledge. 
\textbf{Full CoHa CBOW} is derived from the embeddings trained with CoHa using Word2vec with the CBOW architecture. It consists of the embeddings (i.e., the context matrix) obtained  after the first pass of the CADE approach over the CoHa corpus. It supports the comparison between static and temporal word embeddings trained with a common corpus and algorithm.

\subsection{Metaphor Detection Datasets}
\label{datasets}

\begin{table*}[]
    \centering
     \caption{Dataset Characteristics}
     \resizebox{0.8\textwidth}{!}{
    \begin{tabular}{|c|c|c|c|p{5cm}|}
    \hline
        Dataset & \#sentences & Train/Test Splits  & Temporal Annotation? & Creation Detail \\ \hline
        MOH-X & 646 & No & No & Derived from MOH dataset. The verbs are used as metaphors.  \\ \hline
        VUAsequence & 5323 & Yes  & Yes (1985-1994) & 117 fragments sampled across 4 genres from British National Corpus (academic, news, conversation, and fiction)\\ \hline
        TroFi & 3737 & No & Yes  (1987-1989) &The sentences (each one with a single annotated target verb) are taken from `87-`89 Wall Street Journal Corpus. \\ \hline
    \end{tabular}
    }
\label{tab:dataset}
\end{table*}

Three datasets are used to show the feasibility of the proposed claims.
Table~\ref{tab:dataset} shows the main characteristics of the datasets. 

\textit{\textbf{MOH-X}} \cite{Mohammad} is derived from the subset of MOH dataset used by \cite{shutova3}. Mohammad et al. annotated different senses of WordNet verbs for metaphoricity. They extracted verbs that had between three and ten senses in WordNet along with their glosses. The verbs were annotated for metaphoricity with the help of crowd-sourcing. Ten annotators were recruited for each sentence and only those verbs were selected that were annotated positive for metaphoricity by at least 70\% of the annotators. The final dataset consisted of 647 verb-noun pairs, 316 metaphorical, and 331 literal.

\textit{\textbf{VUA}} consists of 117 fragments sampled across four genres from the British National Corpus, i.e., Academic, News, Conversation, and Fiction \cite{VUA}.
The data was annotated using the MIP-VU procedure \cite{MIP-VU} based on the MIP procedure \cite{discourse}.
The tagset is rich and hierarchically organized, detecting various types of metaphors, words that flag the pre-sense of metaphors, etc.
The majority of sentences in this dataset have the timestamp for the decade 1985-1994.

\textit{\textbf{TroFi}} contains feature lists consisting of the stemmed nouns and verbs in a sentence, with target or seed words. After a first collection phase, the final TroFi dataset is obtained by filtering out some "frequent words" (common words in the British National Corpus along with contractions, single letters, and numbers from 0 to 10).
The target set is built using the `88-`89 Wall Street Journal Corpus (WSJ) tagged using the \cite{Ratnaparkhi} tagger and the \cite{Joshi} SuperTagger.
10-fold cross-validation was adopted on MOH-X and TroFi datasets because of their small sizes (\textit{k} was set equal to 10).
More details regarding the models' hyperparameters can be found in the GitHub repository of \cite{mao-etal-2019-end}'s work. \footnote{\url{https://bit.ly/324ZcUI}}

\section{Experiments}
\label{experiments}
\subsection{Experimental Design}
\label{design}

All data and source code related to our experiments are publicly released.\footnote{\url{https://bit.ly/3qS7NCu}}
A first part of the experimentation consists in evaluating the performance of the RNN HG classifier (evaluated using well-known Precision, Recall, F1-Score and Accuracy measures) when different word embeddings are used instead of GloVe embeddings. In particular, we address the following research questions:
\begin{itemize}
    \item \emph{{\bf RQ1:} Can the results of the state-of-the-art metaphor detection algorithms be improved by using different word embeddings, especially temporal embeddings such as HistWords - SGNS?}
    %obtained with standard GloVe representations (combined with ELMO vectors) 
    \item \emph{{\bf RQ2:} Are there any observable patterns that can lead to the assumption that metaphor detection tasks performed with temporal embeddings and representations impact datasets with known temporal connotations more than others?}
    \item \emph{{\bf RQ3:} Are the representations obtained through Compass alignment more effective for metaphor detection than the embeddings aligned with traditional methods (e.g.: Procrustes)?}
    \item \emph{{\bf RQ4:} Are specific word embeddings' architectures more effective for metaphor detection tasks than others?}
\end{itemize}

The first experiments have been performed using Histwords as new static embeddings. The datasets discussed in Section~\ref{datasets} were used to train a RNN HG model and evaluate it. 
We are interested in evaluating the effectiveness of the new corpus, so we compared the result obtained with HistWords with results obtained using the atemporal Word2vec embeddings trained on the entire Wikipedia corpus.
Finally, experiments have been performed using word embeddings obtained by aligning all different decade slices of the CoHa corpus (ranging from 1820 to 2000) with \textit{Compass} alignment method. Slices of the CoHa corpus for each decade needed to be aligned with Compass in order to perform equivalent experiments to the ones with other embeddings. The first step consisted in concatenating all CoHa text slices, and obtaining a final corpus for all the decades. The pre-processing steps included stripping HTML tags, removing text between square brackets and stop words, and replacing all contractions. The following steps were carried out to perform the alignment using CADE and Compass: (i) Creating the main Compass file by concatenating all the processed CoHa decade slices; (ii) Training the obtained compass-aligned embeddings; (iii) Training all the different slices from the compass obtaining their respective models; (iv) Converting the CoHa compass models in Word2Vec format, so that they could have the same architecture of the HistWords - SGNS and Wikipedia embeddings, and be exploited inside the modified Recurrent Neural Network model.
Only the aligned models of the decade slices ranging from 1900 to 2000 are kept so that the results could be comparable to the previous ones. A \textit{Full CoHa CBOW} (CADE) model was also obtained by training the compass on all the aforementioned decade slices which were used for qualitative analyses.

\paragraph{Qualitative Analysis.}
To get more insights into our results, we also investigate the characteristics of the words that are correctly or mistakenly identified as metaphors, by 1) controlling for linguistic features such as topic and genre, and 2) checking the nearest neighbors of target words in the embeddings used to account for their literal meaning (which account for a more in-depth characterization of word meaning). Due to the large number of experiments performed in this study, we could not inspect all the results. We therefore focused on MOH-X, VUA, and TroFi datasets' predictions obtained with four embeddings: \textbf {(i)} Full CoHa CBOW (CADE); \textbf {(ii)} GloVe (state-of-the-art-representation); \textbf {(iii)} CoHa Word CBOW 1990 Decade Slice; \textbf {(iv)} CoHa Word SGNS 1990 Decade Slice.
Embeddings based on the 1990 decade slice were chosen because of VUA and TroFi datasets' sentences temporal connotations (see Table~\ref{tab:dataset}) and the good performance achieved with these models.
Therefore, the four selected embeddings allowed us to look at predictions made by the RNN HG model with both temporal and atemporal representations and with different embedding and alignment algorithms. In order to check all the predictions made by our model for MOH-X and TroFi, we combined each one of their 10-folds intermediate results, since these two datasets are not split into the train, validation, and test sets like VUA. For each one of the three state-of-the-art datasets, the analysis considered correctly identified metaphors and mistakenly identified metaphors. 
For MOH-X and TroFi only verbs are considered.

    \begin{figure*}[h!]
        \centering
        \includegraphics[scale=0.8]{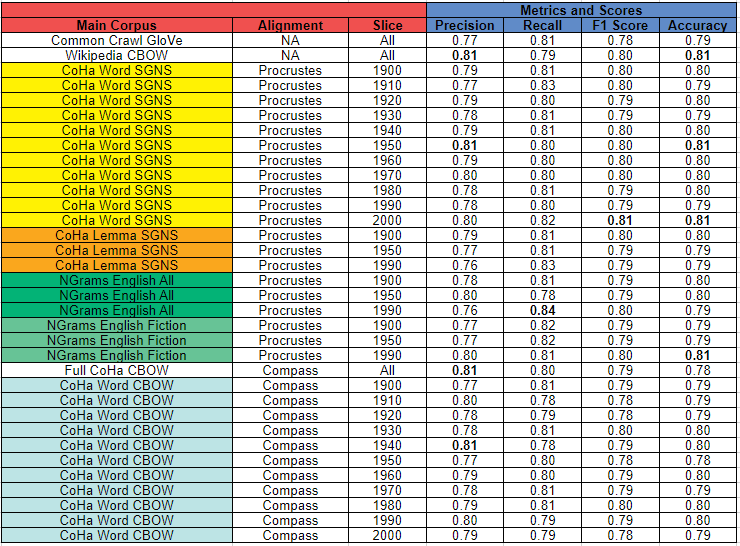}
        \caption{Results related to MOH-X dataset, with every single embedding. \label{im1}}
    \end{figure*}

    \begin{figure*}[h!]
        \centering
        \includegraphics[scale=0.8]{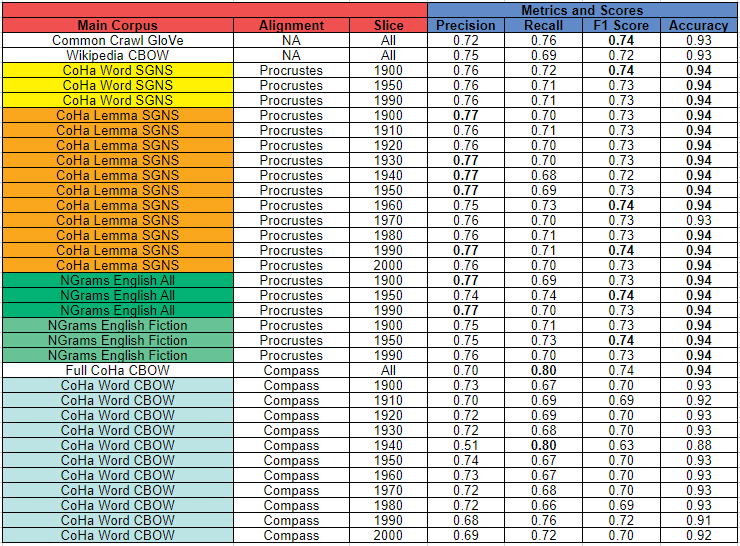}
        \caption{Results related to VUA dataset, with every single embedding. \label{im2}}
    \end{figure*}

    \begin{figure*}[h!]
        \centering
        \includegraphics[scale=0.8]{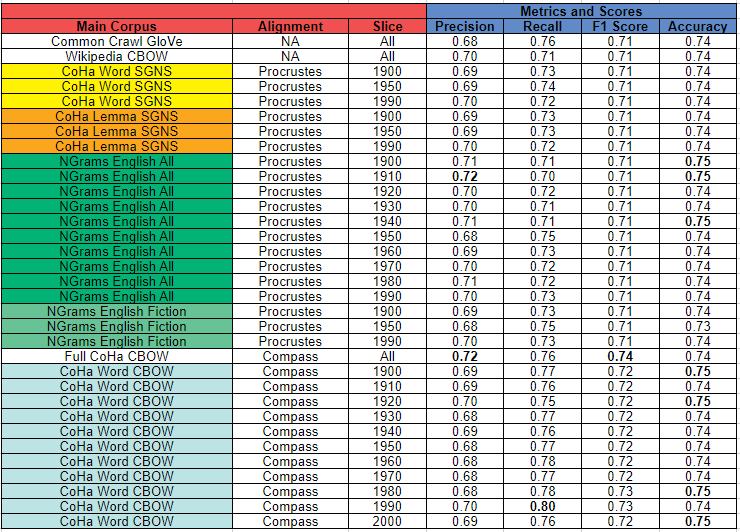}
        \caption{Results related to TroFi dataset, with every single embedding. \label{im3}}
    \end{figure*}
    
\subsection{Quantitative Results}
\label{quantitative}

Tables~\ref{im1},~\ref{im2}, and~\ref{im3} provide the overall quantitative performances and scores for each dataset (the best results are highlighted in bold)\footnote{We only reported a few slices (the most representative ones) for some corpora listed in the tables, such as \textit{Lemma, English All} and \textit{English Fiction}, due to lack of space.}.
Combining the observations gathered from all the performed experiments, the following conclusions can be drawn:
\begin{enumerate}
    \item Word2vec architecture (HistWords - SGNS, CADE, and Wikipedia embeddings) works better than GloVe architecture for metaphor detection;
    \item HistWords - SGNS temporal embeddings perform better on the datasets with known temporal connotations (TroFi and VUA) compared to the atemporal Wikipedia embeddings;
    \item Results on MOH-X are generally good, but they do not show a clear pattern. VUA and TroFi do not show clear patterns either;
    \item Procrustes alignment (HistWords - SGNS) and CADE - Compass alignment methods (CoHa corpus) lead to similar performances and results. Although, while the latter performs better on the TroFi dataset (data extracted from Wall Street Journal corpus), the first one impacts slightly more than the VUA dataset (data extracted from the British National Corpus).
\end{enumerate}

\subsection{Qualitative Results}
\label{qualitative}

This analysis confirmed several expected patterns and revealed some new ones. Among the confirmed patterns, we found that topics related to \textit{economics}, \textit{politics}, and \textit{emotions} are the most recurring ones in sentences containing correctly identified metaphors. Verbs having a literal meaning characterized by physical connotations often assume metaphorical/figurative meanings when used in sentences related to the contexts listed before. This suggests that embeddings derived from these corpora and slices maintain as the core meaning the one related to the physical connotation.  
This pattern is at first observed especially in TroFi predictions, but with the help of the nearest neighbors analysis, the same pattern is detected even in the other datasets. 

The nearest neighbor analysis of the target metaphorical words in the sentences extracted from state-of-the-art datasets leads to comparing meanings of target words in embeddings generated from different resources, especially, in non-temporal vs temporal word embeddings. When exploiting the temporal word embedding, the model could correctly understand that the words (in our examples: ``\textit{apple, virus, attack, hearts} and \textit{glow}") were used in a figurative way, thus correctly classifying them as metaphors. Furthermore, the entire sentences were also correctly classified as metaphorical, since the core meanings of the word were closer to their literal core meaning. This also explains the fluctuations across slices, because corpora are never fully representative, and some contexts may be represented more than others in one specific decade.

Different results related to the domains of the sentences have been observed in the VUA dataset's predictions. With \textit{Full CoHa CADE} embedding, only one sentence belonging to the \textit{academic} genre was correctly classified, whereas as far as \textit{CoHa SGNS 1990 slice} is concerned, no sentences belonging to the \textit{news} genre are correctly predicted. The latter result could indicate that for that specific time period, SGNS words' representations of the \textit{news} genre are biased towards their metaphorical meaning (words are used in metaphorical contexts much more than in literal ones). This would prevent the proposed models from correctly identifying the words as metaphors.

\section{Conclusions}
\label{discussion}
This study can be considered a first attempt to investigate the interaction between metaphorical word usage and semantic change using computational metaphor detection methods and corpus-specific word embeddings, including temporal word embeddings.    

The results suggest that temporal word embeddings can improve the performance of the task of metaphor detection, even though their overall impact on three benchmark datasets is rather limited. However, independently from the absolute performance on the considered datasets, the interaction between the specificity of the embeddings (especially their temporal specificity) and metaphor detection is found in the experiments conducted in this study. In fact, these experiments verify that if the core meaning of the words of interest in a sentence is too similar to their figurative meaning in the word embedding, a metaphorical sentence could get misclassified as literal. Moreover, when temporal word embeddings provide the representations of the words that are more inclined towards their literal core meaning, exploited models end up correctly identifying metaphors more easily. Word embeddings belonging to some language domains in specific time periods can be biased towards their metaphorical meaning, leading to words being used in metaphorical contexts much more than in literal ones. This would prevent neural models from correctly identifying the words as metaphors.
To improve the experimental framework, both temporal and atemporal representations could be built on the same corpora with temporal slices. Furthermore, when building atemporal embeddings, the corpora could be subsampled to obtain a comparable size.

Future work may stem from our last exploratory analysis. Searching for words known to undergo semantic change across time, we retrieved a suitable list from the \textit{SemEval 2020 Task 1: Unsupervised Lexical Semantic Change Detection Competition}\footnote{\url{https://bit.ly/3Gz9vPU}}. We searched for all the occurrences of these words in the three datasets used in our study and in the competition data, to classify the word usage as metaphorical or not, but we could not find enough metaphorical statements. This suggests that more work is needed to collect more data better accounting for the interaction between metaphorical word usage and semantic change along time, a phenomenon that is advocated by many scholars as a very important driver of language evolution.

\bibliographystyle{unsrtnat}
\bibliography{references}  %%% Uncomment this line and comment out the ``thebibliography'' section below to use the external .bib file (using bibtex) .





% \end{thebibliography}

\end{document}